# Women Sport Actions Dataset for Visual Classification Using Small-Scale Training Data

Palash Ray[1], Mahuya Sasmal[1], and Asish Bera[2, *]

[1]Department of Computer Science and Engineering, Haldia Institute of Technology, WB, India

[2]Department of Computer Science and Information Systems, BITS Pilani, Pilani Campus, Rajasthan 333031, India.

*Corresponding author: asish.bera@pilani.bits-pilani.ac.in

**Abstract**

Sports action classification representing complex body postures and player-object interactions is an emerging area in image-based sports analysis. Some works have contributed to automated sports action recognition using machine learning techniques over the past decades. However, sufficient image datasets representing women's sports actions with enough intra- and inter-class variations are not available to the researchers. To overcome this limitation, this work presents a new dataset named WomenSports for women's sports classification using small-scale training data. This dataset includes a variety of sports activities, covering wide variations in movements, environments, and interactions among players. In addition, this study proposes a convolutional neural network (CNN) for deep feature extraction. A channel attention scheme upon local contextual regions is applied to refine and enhance feature representation. The experiments are carried out on three different sports datasets and one dance dataset for generalizing the proposed algorithm, and the performances on these datasets are noteworthy. The deep learning method achieves 89.15% top-1 classification accuracy using ResNet-50 on the proposed WomenSports dataset, which is publicly available for research at Mendeley Data.

## Introduction

Sports activities reflect intricate human-object interactions and body-part movements, as well as integrity and diversity across the world [1]. Computer vision and machine learning methods are widely used in several domains of sports, *e.g.*, activity classification, complex human-object interaction, object detection, player recognition, motion analysis, ball tracking, etc. [2], [3], [4], [5], [6]. The analysis of diverse sport actions is crucial in several international competitions, *e.g.*, Olympic, Commonwealth, World Cup, etc., for automated motion and posture analysis, ball tracking, foul prediction, player recognition, etc. Sport actions represent complex body-pose movements and object interactions under variable lighting conditions and intricate background [7], [8]. Over the past decades, several automated systems for sports analysis have been developed using artificial intelligence (AI) and machine learning (ML) techniques. However, little research attention has been focused on sport applications using deep learning (DL) methods that require large amounts of data for training [9], and only a few sports video datasets are publicly available for research (Table 1). Most of these datasets demonstrate the male players. Particularly, the sporting actions of women athletes have not been widely explored in existing literature. It is our main aim to develop a new dataset consisting of different sports actions of women players, named **WomenSports**. Another objective is to devise a deep learning approach for sports actions classification with limited training data. The proposed dataset will be explored further in the context of sports for developing enhanced state-of-the-art technology. Also, it bridges the gap of realistic data availability for small-scale training-based ML techniques, where data scarcity is a challenge for women's sports. Thus, the WomenSports dataset widens other exploratory areas of sports. This new dataset will be helpful for research for several reasons:

- There are insufficient image datasets on sports, which provide labeled data exclusively for different sports activities performed by women.
- The WomenSports dataset can be utilized by sports organizations, coaches, and players to improve performance analysis in women's sports. The dataset enables the evaluation of new algorithms that can precisely identify and interpret different motions and movements, leading to a detailed comprehension of sports performance.
- The dataset can be used as a basis for creating sports-related technology, including automatic event recognition systems, real-time performance monitoring tools, and interactive sports applications, tailored specifically for women sports. The dataset will encourage researchers in the fields of artificial intelligence, sports engineering, and data analytics.

Convolutional neural networks (CNNs) have contributed significantly to discriminate visual variability in several image recognition tasks, including sport actions [10], tracking sport data and analysis [11], [12], [13], etc. Sample examples of such relevant areas are shown in Fig.1. CNNs effectively aggregate complex patterns and spatial relationships in the feature space. A CNN extracts deep features through successive non-linear activations by transforming an input image describing low-level information into a high-level feature vector. Several state-of-the-art standard CNN backbone families are available, such as ResNet, Inception, MobileNet, and others. These CNNs perform better than traditional feature descriptors, even with limited and small-scale datasets. Moreover, existing works show a pathway of learning local details at multiple contexts which aggregate essential information into an efficient feature map. In this regard, the attention mechanism is very effective in localizing the most discriminatory regions within an input image. Here, learning vital local description, which is further enriched by channel attention, is studied to enhance overall high-level feature representation. Data augmentation is an effective method for generating wide variations in the input data during training, which in turn improves learning efficacy [8]. Data augmentation randomly imposes diverse geometric and spatial transformations within training samples for data diversity when training data is limited and small.

The contribution of the proposed deep learning model is an aggregation of feature maps computed from multiple patches using standard backbone CNNs. The experiments are mainly carried out on the WomenSports dataset. Other public datasets of sports and dance poses are evaluated to generalize the efficacy of the proposed algorithm. The main contributions of this paper are:

- Aggregation of region-based local feature maps enriched with an attention mechanism for sports action recognition.
- Development of a new image dataset consisting of 50 categories of women's sports with class-label annotations for public research especially, constrained with limited training data.
- Experiments are conducted on the WomenSports-50 dataset. In addition, experimental analysis on other sports and dance datasets are provided for generalizing the proposed work.

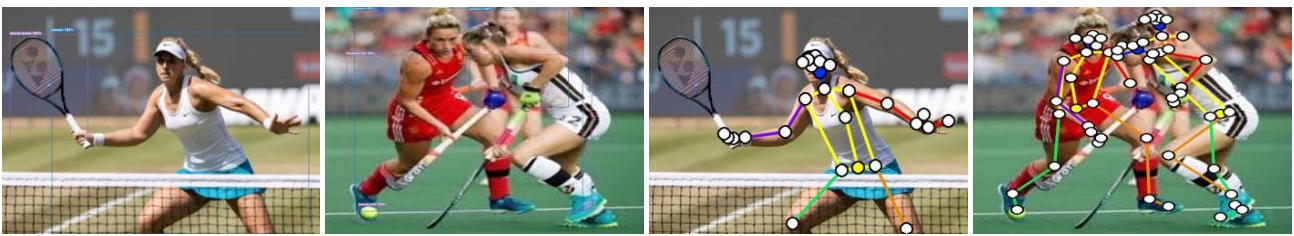

**Figure 1.** Possible applications of the WomenSports dataset in visual classification, object detection, and human pose-estimations using body-joints and key-points localization.

**Table 1.**
Summary of a few published image/video datasets on sports recognition.

| Reference, Year | Dataset Description |
|---|---|
| [14], 2010 | Sports 8: 1000 training and 1000 testing images. |
| [15], 2011 | UCF Sports: 150 clips at 10 fps |
| [16], 2014 | 1 million YouTube videos, 487 sports classes |
| [17], 2021 | Basketball-51: comprises a total of 10311 video clips categorized into 8 classes, recorded from 51 NBA basketball games |
| [8], 2023 | Sports 102: General sport action images of 102 classes |
| Proposed, 2025 | **WomenSports**: Women sport actions with limited training data of fifty classes |

**Related Works**

Visual classification of sports and related human actions using deep learning have witnessed a significant improvement in performances compared to traditional ML techniques. A brief study on a few publicly available image/video sports datasets is presented in Table 1.

The Leeds Sports Pose (LSP) dataset was developed in the earlier decade [14]. A deep network built with long short-term memory (LSTM) units was tested to benchmark the LSP dataset. The suitability of computer vision applications like ball and players detection and tracking including motion analysis was explored [18]. A two-stream attention model using LSTM was presented for action recognition [19]. A method for athlete's incorrect action recognition based on a dual-channel 3D CNN including spatial attention was developed [20]. The foul actions of the Chinese basketball players based on machine vision technique and Gaussian model method were tested [21]. Recently, a system for foul behavior detection in soccer games has been developed by integrating an IoT (internet of things) system with a YOLOv8 model exploiting residual local feature networks, called VAR-YOLOv8s [22]. A multi-scale feature fusion model was developed using a 3D attention mechanism for fine-grained basketball motion recognition [23]. A deep neural network was fused with a tuna swarm optimization (NTSO) for sports image classification [24]. An ensemble neural network was developed for sport actions (squats, pull-ups, and dips) classification using the data collected by inertial sensors [25]. A time-weighted motion history image was introduced for generating frame-level importance in the context of sport motion analysis with small-scale data [7]. Also, an automated low-cost umpiring system based on a multi-view ball tracking technique for table tennis was developed [6].

A fusion-based global and local pose information refinement network was proposed [26]. The SYD-Net was developed using an attention mechanism for recognizing diverse sport actions, yoga poses, and dance postures [8]. The SYD dataset was tested using a *max-min* normalization at preprocessing, and then YOLOv5 with *k*-means clustering, and coordination attention [27]. Also, YOLOv4 was utilized to extract informative features from the basketball videos, and for classification, LSTM and type-2 fuzzy logic were used [28]. Collective Sports, called the C-Sports dataset, represents a multi-task recognition of both collective activity and sports categories. A CNN with an LSTM model was used for evaluating the C-Sports dataset [29]. A motion history sequential representation was proposed for action recognition [7].

A few video-based sports datasets were developed in the literature, such as the UCF Sports dataset [30] containing 150 short videos. A global scene model was tested on this dataset [15]. The spatio-temporal features were computed from YouTube videos consisting of 487 sports classes, *e.g.*, football, running, cricket, and others [16]. The DeepSportradar-v1 dataset was developed for automated sport understanding [31].

The SportsMOT represents a multi-object tracking video-based dataset in diverse sports scenes [32]. In contrast, the SportsPose described a human 3D pose dataset consisting of dynamic sports movements [33]. The MultiSports dataset was enriched with dense spatio-temporal actions of four sports classes [34]. An activity recognition technique was proposed for fine-grained sports recognition using videos [35]. A pictorial structural model was developed with clusters of pose and appearance features computed from different sport activities and evaluated with a mixture of classifiers [36]. Another pictorial structure model was tested with mixtures of trees [14].

A deep learning method integrated with CNNs, and graph convolutional networks (GCNs) was presented to learn intricate spatio-temporal patterns for football player action recognition [37]. The SoccerAct dataset contains 3557 YouTube videos of ten soccer actions, like free kick, penalty shot, etc. [38]. Drone-based videos of various games, such as tennis, and soccer, were evaluated [39]. Using CNN, players were detected in field sports too [40]. Automated analysis of a catch attempt in American football based on audio-visual data was studied [41].

Another direction of research includes few-shot learning and domain adaptation techniques for human action and sports actions recognition [42], [43]. In few-shot learning, only a few samples (*e.g.*, 5-way) per class are trained, while tested in unseen classes. These methods exhibit a more challenging scenario for image and video classification. The ProtoGAN describes a conditional generative adversarial network for a few-shot learning protocol to classify video-based sport actions [44]. Recently, an unsupervised domain adaptation approach has been introduced by minimizing interdomain and intra-domain discrepancies.

The method developed a recurrent neural network exploiting LSTMs for video-based action classification [45].

Most of these approaches have studied diverse sports using images and videos, played by both males and females. This current study contributes a new dataset containing 50 sports categories performed by women players constrained with limited/small-sized training data. In contrast to conventional few-shot learning that requires unseen classes for generalization, here, we follow a standard classification method. In our scheme, we have trained with all 50 known classes, but limited to only $k$-samples per class, like [42]. This kind of small training set of diverse sport categories spurs new challenges to the researchers.

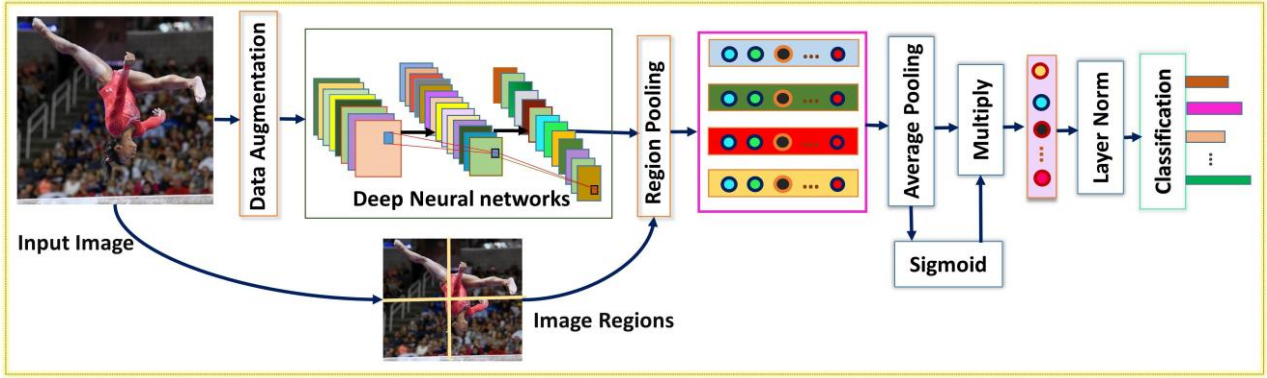

**Figure 2.** A proposed deep learning model was developed for WomenSports action recognition.

**Algorithm 1** Proposed Algorithm

**Input:** Image $I$, Number of regions $R$, learning parameter: $\theta$

**Output:** predicted class-label $Y_{pred}$ of human action

1. $I' \leftarrow ImageAugment(I)$
2. **Initialize** a base CNN with the learnable parameters $\theta$

**repeat**

   3. $F \leftarrow DeepConvoultionalNetwork(I')$

   4. $F_{up} \leftarrow Upsampler(F)$

   5. $F_r \leftarrow RregionPooling(F_{up}, R)$

   6. $F_p \leftarrow AveragePooling(F_r)$

   7. $F_A \leftarrow F_p \otimes Sigmoid(F_p)$

   8. $F_{final} \leftarrow LayerNorm(Dropout(F_A))$

   9. $Y_{pred} \leftarrow Softmax(F_{final})$

   10. compute loss $L_{ce}$ and optimize parameters $\theta$;

**until** $\theta$ converges

**end**

## Methodology

A DL method is developed for aggregating high-level features demonstrating the subtle distinctness of diverse sport actions. The proposed method is briefed in Algorithm-1 and shown in Fig. 2.

First, the training samples are preprocessed and augmented to incorporate random variations by common image augmentation techniques such as rotation, scaling, and blurring. An input image with class-label $I_l \in R^{h \times w \times 3}$ is passed into a base convolutional neural network (CNN) which computes a deep feature map $F \in R^{h \times w \times c}$ where $h$, $w$, and $c$ imply the height, width, and channels, respectively. The base output feature map is enhanced for capturing crucial information through successive layers, as described next.

### Region Pooling

Region-based local feature descriptions have shown promising success in learning local contexts, which are represented as stacked layers of fixed-size feature descriptors [50], [52]. To enrich local representational efficacy, a base feature map is divided into four non-overlapping local regions. The base feature map, having a lower spatiality, is upsampled to 32×32 resolution for encoding the usefulness of local contexts at a bigger scale. Then, four regions ($N$) are defined with the encoded feature map, and a fixed-dimensional feature map is pooled from each region. The pooled feature vector is denoted as $F \in R^{N \times h \times w \times c}$, where $N$ is the number of regions. Each region represents either a horizontal slice with (h×w/2) or (h/2×w) size of the upscaled and encoded feature map.

$$F_{pool} = GAP\ (RegionPooling\ (F)) \quad (1)$$

### Channel Attention Mechanism

Here, channel attention is inspired by the widely used convolutional block attention module (CBAM) that disentangles channel-wise feature space and squeezes global spatial information using a global average pooling (GAP) layer [46]. Also, this attention scheme can be related to the squeeze operation, excluding fully connected layers, as followed in squeeze-and-excitation network (SENet) [47]. CBAM carries expressive channel-wise statistics of global summary. A channel attention map is generated by exploring the relationship of inter-channel features and is effective in selecting significant features. The channel-wise attention plays a vital role in focusing on 'what' is informative in region-pooled feature maps.

For this intent, spatial dimension of region-pooled feature maps is squeezed precisely ($h \times w \times c \rightarrow 1 \times 1 \times c$) for aggregating structural information through a GAP layer that renders a mean channel-wise descriptor across channel dimension, denoted as $F_{GAP} \in R^{N \times 1 \times 1 \times c}$. Next, channel-wise features are converted into an attention map by exploiting nonlinearity with a *sigmoid*(σ) activation in the pooled feature map $F_{pool}$ and multiplied elementwise with the actual pooled features. This attention method recalibrates the crucial channel features for the classification of diverse sports categories. This attention scheme does not contribute to any additional computational parameter yet carries a relevant information summary as a collective local descriptor of the whole image. These important stages are specified in Algorithm 1, and the equations are given next.

$$F_{attention} = Sigmoid\ (F_{pool}) \otimes F_{pool} \quad (2)$$

$$\hat{F} = LayerNorm(Dropout(F_{attention})) \quad (3)$$

$$Y_{pred} = softmax(\hat{F}) \quad (4)$$

### Classification

Lastly, a feature vector $\hat{F}$ is obtained by passing through layer normalization and drop-out layers successively. Then, $\hat{F}$ is fed to a *softmax* layer for generating an output predicted probability of each class-label $\bar{l}$ corresponds to original class-label $l \in Y$ from a collection of classes $Y$. The cross-entropy loss is used for multi-class classification of sport actions. If there are $Y$ classes and $y_l$ represents the true label of class $l$, and $p_l$ implies the predicted probability implying class $l$, then categorical cross-entropy loss $L_{CE}$ is defined as:

$$L_{CE} = -\sum_{l=1}^{Y} y_l \log(p_l) \quad (5)$$

The $L_{CE}$ calculates the discrepancy between the true distribution ($y_l \in Y$) and the predicted distribution ($p_l$) of class labels.

Layer normalization and dropout layers are included to mitigate overfitting issues during the learning phase. The Stochastic Gradient Descent (SGD) optimizer is applied for batch-wise training and computing the categorical cross-entropy loss (Algorithm 1: steps 8-9). The SGD is chosen for two reasons. Firstly, SGD reduces the variance during parameter update and provides a stable and faster convergence. Secondly, it is computationally advantageous due to its capacity of optimized matrix operations and gradient flow. Also, randomness in SGD algorithm is useful to overcome local minima and assist to reach global minima. In addition, it is highly adept while training on large datasets. Here, SGD optimizer is chosen to leverage these benefits.

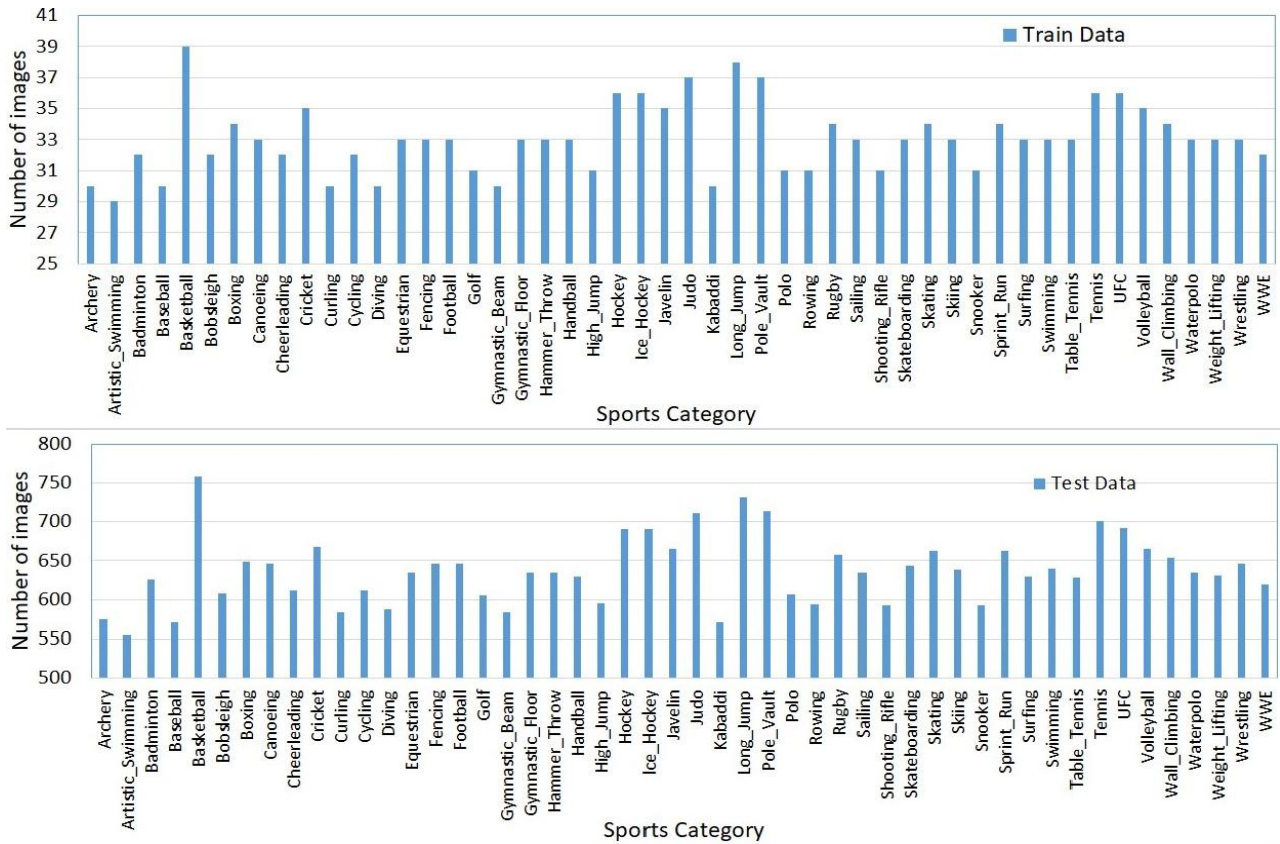

**Figure 3.** Class-wise distribution of the WomenSports limited-training (5%) and test (95%) images.

## WomenSports Dataset

The WomenSports dataset depicts dynamic moments in different sports. This dataset is intended for visual classification, specifically for identifying and categorizing sport actions performed by female athletes. The dataset is built from many sources, including popular sporting event websites, stock picture libraries, and several public photographic archives. The sports categories are selected based on their global popularity and representation in women's sports events. A balanced mix of indoor and outdoor sports are included to ensure diversity and comprehensive coverage. Popular sports like basketball and football are included alongside niche sports like fencing to create a representative dataset. There are 50 sports classes, including basketball, soccer, tennis, swimming, gymnastics, volleyball, and others (detailed in the supplementary document). The purpose of each image is to capture vital moments of sports. These moments include athletes dribbling a basketball, rushing on the field, doing dives, striking a tennis ball, and other similar activities. The images are annotated manually by a team of three domain experts with a background in sports and image processing. Each expert independently labeled the images, and discrepancies have been resolved through a consensus-based approach that ensures the consistency of annotations. We have employed a multi-step labeling process: initially, images are categorized by sport based on visible cues such as player uniforms, equipment, and the environment (indoor or outdoor). Next, each image is reviewed for correctness and labeled with a unique identifier corresponding to its category. Finally, quality checking is performed to validate the labels.

During the dataset preparation stage, data cleaning involved removing duplicate, low-quality, or ambiguous images that could compromise classification performance. Our team assessed ambiguity and retained only images with clear and distinct visual cues. The images are standardized such that they are adequate for visual categorization using class labels. These source websites already provide class labels for most of the samples, such as Kaggle's datasets. However, there are many duplicate, noisy, and mislabeled samples, which are corrected by a team of student volunteers who have sufficient skills to identify all distinct sport classes under the supervision of domain experts at academic institutes. They have assisted in developing this dataset for research. With this rigorous annotated dataset, fine-grained sport activities categorization is performed.

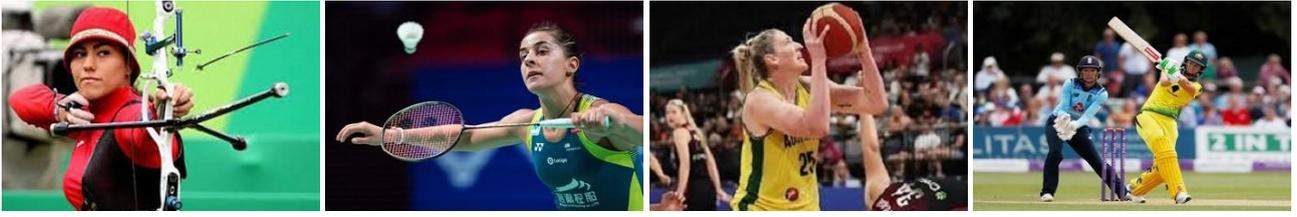

**(a)** Archery, Badminton, Basketball, Cricket

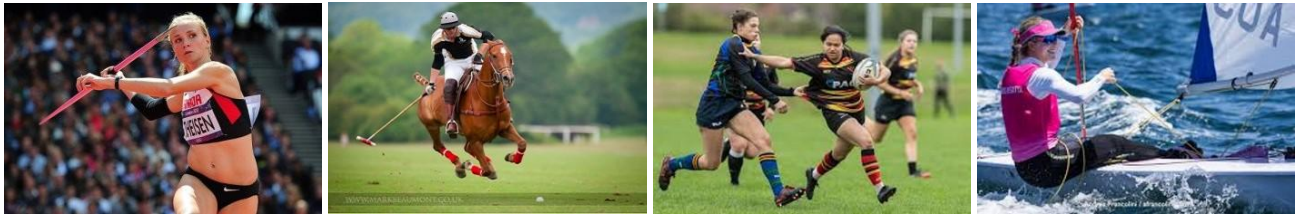

**(b)** Javelin, Polo, Rugby, Sailing

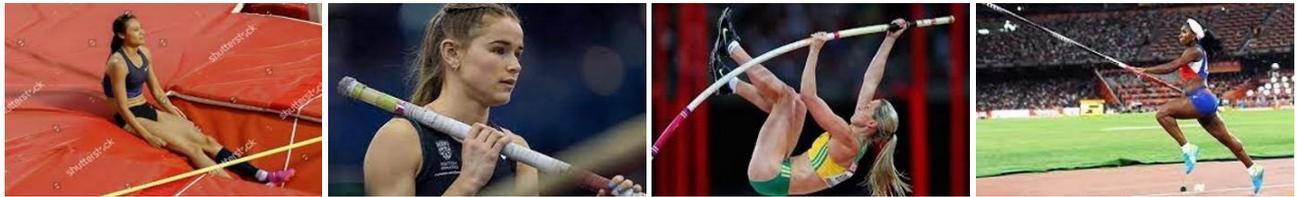

**(c)** Pole Vault

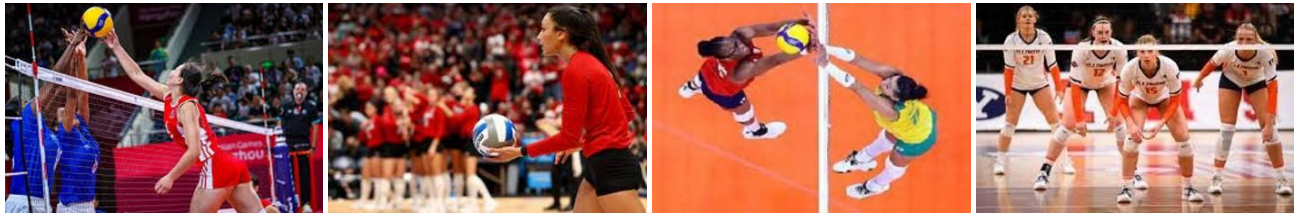

**(d)** Volleyball

**Figure 4.** Samples of the WomenSports dataset showing (a-b) inter-class and (c-d) intra-class variations.

## Description of the WomenSports Dataset

The WomenSports-50 dataset contains 33504 images, and the training and test sets are shown in Fig. 3. Fig. 4 shows a few samples of this dataset. The training set comprises 5% of the total dataset, implying 1676 samples. The main reason for data distribution is the evaluation protocol.

Conventional deep learning-based image classification approaches require huge, labelled training data, which is often a challenge in practical scenarios. To address high data requisites and label annotations, several strategies like active learning and few-shot learning methods have been developed in the literature [48]. In prior works, several image recognition benchmark datasets have been split with a larger (or almost equal) size of test-set than the training-set size. For example, fine-grained image recognition, remote sensing scene classification, medical image classification, and others [49]. Inspired by such conditions, we formulate this evaluation protocol considering that the training data size is small, *i.e.*, only 5% of the total dataset size. It is our main objective to study how, with a small amount of training and validation data samples, the performance of a deep learning model can be improved. However, other ratios of samples can also be evaluated. Hence, we pose the current image classification problem with limited training (5%) and validation (5%) data and tested with the remaining 90% samples. The efficiency of the proposed model is improved over the baseline results. The proposed dataset distribution will encourage the researchers to develop more powerful methods and improve the results further soon. The dataset will be studied further for object detection, sports analysis, and related applications.

### *Description of other Sports and Dance datasets*

In addition to the proposed women sports dataset, other publicly available datasets on sport and dance actions are evaluated for analyzing the effectiveness of the proposed method. A brief description of these datasets representing image distributions is summarized in Table 2. The Kaggle dataset is collected and prepared for training with limited samples, like the proposed women's sports dataset. The Sports-100 dataset contains only 500 samples for training and validation, each in our experiments. The Sports-102 [8] and Dance-20 [3] datasets are recently published and experimented for performance comparison with the proposed work.

## Experimental Results

Details about the implementation strategy and experimental study are provided. Next, an ablation study is presented, and sample visualizations are shown.

### *Implementation Specifications*

Pre-trained CNNs, namely, the Inception-V3, Xception, ResNet-50, and MobileNet-V2, are used for convolutional feature computation. The Inception module is decoupled into channel-wise and spatial correlations by pointwise and depth-wise separable convolutions, which are building blocks of the Xception architecture. Separable convolution follows depth-wise convolution for spatial and pointwise convolution for cross-channel aggregation into a single feature map. The ResNet architecture uses a residual connection and represents an identity mapping through a shortcut connection following the simple addition of feature maps of previous layers. The MobileNet-V2 uses bottleneck separable convolutions and inverted residual connections. It is a memory-efficient framework suitable for mobile devices. These backbones are widely used due to their superior architectural design at a reasonable computational cost. The model parameters of CNNs are estimated in millions (M).

Basic pre-processing methods offered by the Keras applications are applied. Converting the input images from RGB to BGR is required. Then, each color channel is zero-centered with respect to the ImageNet dataset, without scaling. The proposed method is developed in Tensorflow 2.x using Python. The versions are as follows:

- Tensorflow: 2.13.0, Keras: 2.13.1, Cuda: 12.4, NVIDIA A100 40GB GPU.

- Intel Core Silver 4316 CPU x86 64, 2.30 GHz 128GB RAM.

**Table 2.** Distributions of train, validation, and test sets of datasets.

| Reference | Dataset | Class | Train | Validation | Test |
|---|---|---|---|---|---|
| Proposed | Women Sports | 50 | 1676 | 1676 | 30152 |
| Kaggle [51] | Sports 100 | 100 | 500 | 500 | 13493 |
| SYD-Net [8] | Sports 102 | 102 | 9278 | - | 4320 |
| POA-Net [3] | Dance | 20 | 5565 | - | 2760 |

**Table 3.** Baseline performances (%) on the WomenSports dataset.

| Backbone CNN | Top-1 Acc | Top-3 Acc | Precision | Recall | F1-score | Param (M) |
|---|---|---|---|---|---|---|
| ResNet-50 | 76.47 | 89.73 | 77.0 | 76.0 | 76.0 | 23.7 |
| Xception | 73.40 | 88.36 | 74.0 | 73.0 | 73.0 | 21.0 |
| Inception-V3 | 70.19 | 85.68 | 71.0 | 70.0 | 70.0 | 21.9 |
| MobileNet-V2 | 73.71 | 88.20 | 74.0 | 74.0 | 74.0 | 2.3 |

**Table 4.** Performance (%) of the proposed method with Gaussian blur data augmentation and regularization on the WomenSports dataset using various base CNNs.

| Backbone CNN | Top-1 Acc | Top-3 Acc | Precision | Recall | F1-score | Param (M) |
|---|---|---|---|---|---|---|
| ResNet-50 | 80.16 | 91.96 | 80.0 | 80.0 | 80.0 | 23.7 |
| Xception | 78.32 | 91.37 | 79.0 | 78.0 | 78.0 | 21.0 |
| Inception-V3 | 77.13 | 90.80 | 78.0 | 77.0 | 77.0 | 21.9 |
| MobileNet-V2 | 75.75 | 88.65 | 76.0 | 75.0 | 75.0 | 2.3 |

**Table 5.** Performance (%) on the WomenSports dataset using the proposed method.

| Backbone CNN | Top-1 Acc | Top-3 Acc | Precision | Recall | F1-score | Param (M) |
|---|---|---|---|---|---|---|
| ResNet-50 | 89.15 | 97.00 | 92.0 | 90.0 | 91.0 | 27.9 |
| Xception | 88.70 | 96.88 | 89.0 | 89.0 | 89.0 | 25.2 |
| Inception-V3 | 88.33 | 96.03 | 88.0 | 88.0 | 88.0 | 26.1 |
| MobileNet-V2 | 88.22 | 95.13 | 88.0 | 88.0 | 88.0 | 4.0 |

**Table 6.** Baseline top-1 accuracy (%) on (a) Sports-100 dataset with 5% train, 5% validation, and 90% test samples, and (b) Dance-20 dataset.

| Dataset | ResNet-50 | Xception | InceptionV3 | MobileNetV2 |
|---|---|---|---|---|
| Sports-100 | 68.55 | 68.14 | 64.11 | 64.51 |
| Dance-20 | 81.43 | 79.90 | 77.83 | 77.21 |

**Table 7.** Performance (%) on the WomenSports dataset with 5% train, 5% validation, and 90% test ratio using proposed method.

| Dataset | Backbone CNN | Top 1 Acc | Top 3 Acc | Precision | Recall | F1-score |
|---|---|---|---|---|---|---|
| WSports-50 | ResNet-50 | 90.16 | 93.31 | 90.0 | 90.0 | 90.0 |
| | Xception | 90.90 | 93.56 | 91.0 | 91.0 | 91.0 |
| | Inception-V3 | 88.70 | 92.56 | 93.0 | 91.0 | 92.0 |
| | MobileNet-V2 | 88.35 | 87.54 | 89.0 | 87.0 | 88.0 |
| Sports-100 | ResNet-50 | 78.68 | 90.51 | 78.0 | 76.0 | 77.0 |
| | Xception | 79.07 | 92.54 | 80.0 | 78.0 | 79.0 |
| | Inception-V3 | 75.56 | 90.25 | 77.0 | 75.0 | 76.0 |
| | MobileNet-V2 | 78.12 | 92.31 | 80.0 | 78.0 | 79.0 |

CNNs are trained with a 256×256 image size. For data augmentation, rotation (±25 degrees), zooming (±0.25), and cropping with 224×224 image-size are applied. The deep learning models are trained for 100 epochs, starting with a 0.001 learning rate.

$$Accuracy = \frac{TP+TN}{TP+TN+FP+FN}, \quad Precision = \frac{TP}{TP+FP}$$

$$Recall = \frac{TP}{TP+FN} \quad F1-Score = 2 \times \frac{Precision \times Recall}{Precision+Recal}$$

(6)

where *TP* is the number of true positives, *TN* is the number of true negatives, *FP* is the number of false positives, and *FN* denotes the number of false negatives.

*Result Analysis*

Performance is evaluated using standard metrics: top-1 accuracy, top-3 accuracy, average precision, recall, and F1-score (eq. 6). These metrics are widely used to evaluate predictive performance when the samples of various classes are imbalanced. In addition, we have evaluated performance using a confusion matrix, which provides a reliable performance assessment of our model.

Table 3 shows the baseline performances of backbone CNNs. The parameters of deep learning models are given in millions (M) in the last column. Pre-trained ImageNet weights are used to initialize backbone CNNs, and the common data augmentation techniques are considered. The base feature maps of base CNNs are passed through a GAP and *softmax* layer. However, no regularization method has been used in the baseline evaluation. ResNet-50 attains the highest 76.47% top-1 accuracy on the WomenSports dataset.

Next, the Gaussian blur and random flip are used as additional data augmentation, included with other transformations, which are used in baseline evaluation. Also, dropout and layer normalization are included to ease overfitting. The results are given in Table 4. The results imply that there are improvements in performance using four backbones without adding additional parameters. ResNet-50 attains the highest 80.16% top-1 accuracy on the WomenSports dataset. Thus, data augmentation is very effective in producing more data variations when training data is limited, which is tackled here.

Table 5 gives the performances of the proposed method, including a summary of local contextual information and refined with channel attention. The results imply that the current method has boosted performance for classifying sports actions while trained with limited data. The highest accuracy of ResNet-50-based model is 89.15%, which is a good margin compared to the respective baseline accuracy of 80.16%. Likewise, gains of other backbones are noteworthy.

In another setting, with training 5%, validation 5%, and testing, a 90% data split has been maintained for conducting experiments with the same deep learning method. The results of the baseline evaluation and the full proposed model are given in Tables 6 and 7, respectively. The results are improved marginally using different backbone CNNs. A reason could be the use of a validation set which in turn reduces test-set data. The validation set provides support to the model for improving learning effectiveness during training phase.

The performances are evaluated on other datasets, particularly Sports-100, Sports-102, and Dance-20, for demonstrating the generalization ability of the proposed method, given in Table 8. The highest top-1 accuracy on the Sports-100 is 79.07%, achieved by Xception. Likewise, the same backbone attains the best top-1 accuracy of 96.66% on the Sports-102 and 93.75% on Dance-20 dataset. The model parameters of various CNNs differ due to their architectural design. The proposed model is built with 27.9M parameters using ResNet-50. In contrast, the lightweight MobileNet-V2-based model requires only 4.0M parameters. Still its performances are competitive with other standard backbones.

The supplementary document shows the confusion matrices evaluated by the proposed method using different backbones. The confusion matrix using MobileNet-V2 on the Dance-20 is shown in the supplementary document (Fig. 8).

**Table 8.** Performance on other human activity recognition dataset.

| Dataset | Backbone CNN | Top-1 Acc | Top-3 Acc | Precision | Recall | F1-score |
|---|---|---|---|---|---|---|
| Sports-102 | ResNet-50 | 95.92 | 99.27 | 96.0 | 96.0 | 96.0 |
|  | Xception | 96.66 | 99.28 | 97.0 | 97.0 | 97.0 |
|  | InceptionV3 | 96.50 | 99.23 | 96.0 | 96.0 | 96.0 |
|  | MobileNetV2 | 96.10 | 99.12 | 96.0 | 96.0 | 96.0 |
| Dance-20 | ResNet-50 | 91.90 | 99.22 | 92.0 | 92.0 | 92.0 |
|  | Xception | 93.75 | 99.24 | 94.0 | 94.0 | 94.0 |
|  | InceptionV3 | 91.38 | 99.44 | 92.0 | 91.0 | 91.5 |
|  | MobileNetV2 | 91.19 | 99.88 | 91.0 | 91.0 | 91.0 |

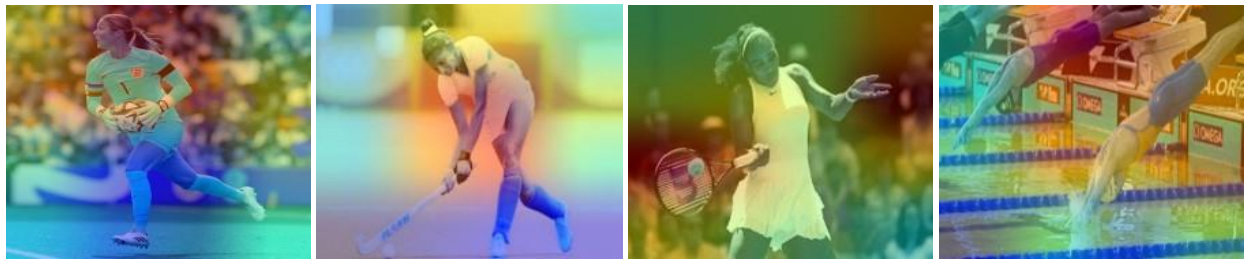

**(a)** inter-class: football, hockey, tennis, and swimming

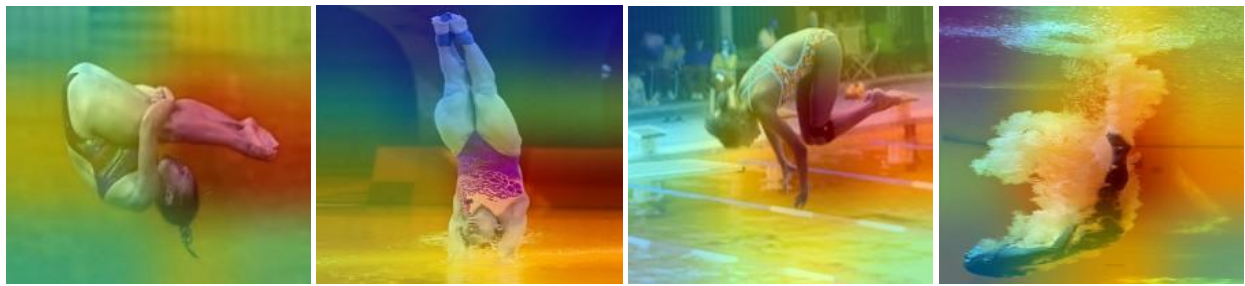

**(b)** intra-class: diving actions

**Figure 5.** Grad-CAM visualizations illustrate coarse image-regions which are suitable for making predictions of various inter-class and diversified intra-class sport actions.

**Table 9.** Ablation study: performance of regions upon baseline (backbone CNNs) on the WomenSports dataset.

| Backbone CNN | Top-1 Acc | Top-3 Acc | Precision | Recall | F1-score |
|---|---|---|---|---|---|
| ResNet-50 | 88.10 | 94.20 | 94.0 | 94.0 | 94.0 |
| Xception | 86.68 | 91.24 | 88.0 | 86.0 | 87.0 |
| InceptionV3 | 87.84 | 95.79 | 89.0 | 87.0 | 88.0 |
| MobileNetV2 | 86.70 | 91.03 | 87.0 | 87.0 | 87.0 |

**Table 10.** Ablation study: performance with channel attention added upon baseline (base CNNs) on WomenSports dataset.

| Backbone CNN | Top-1 Acc | Top-3 Acc | Precision | Recall | F1-score |
|---|---|---|---|---|---|
| ResNet-50 | 81.88 | 93.09 | 82.0 | 82.0 | 82.0 |
| Xception | 78.92 | 91.22 | 79.0 | 79.0 | 79.0 |
| InceptionV3 | 77.94 | 90.07 | 78.0 | 78.0 | 78.0 |
| MobileNetV2 | 77.81 | 89.76 | 78.0 | 78.0 | 78.0 |

*Performance Comparison*

**WSports-50 and Sports-100:**

This work introduces a new evaluation strategy while training with limited samples of the WSports-50 and Sports-100 datasets. This method's performances are not directly comparable with existing works. However, a comparative study is given to complete the discussion in the present work.

With a different image distribution of train-test split, work on a similar women's sports dataset has achieved 80.16% accuracy [9]. Using conventional image distribution of Sports-100, a squeeze and excitation network upon a pre-trained residual network, called SE-RES-CNN, has achieved 98.0% accuracy on this dataset [51]. Here, we have formulated a new challenge with limited training and validation samples, and our method benchmarks 78.68% accuracy on Sports-100 using ResNet-50 base. Likewise, the accuracy on WSports-50 is 90.16% while evaluated with only 5% training, 5% validation, and 90% test samples using ResNet-50. The best results are achieved using Xception. The results are given in Table 7.

**Sports-102**: SYD-Net with 9 regions has reported 95.66% accuracy using Xception base while trained from scratch, and with pre-trained *ImageNet* weights, their accuracy is 98.81% with 20 regions [8]. In contrast, this work uses only four regions over Xception and achieves 96.66% accuracy, which is a decent performance regarding the model complexity (Table 8). The proposed work attains 96.66% accuracy on Sports-102, which is competitive with SYD-Net's 96.70% with 30 patches using Xception. SYD-Net is a comparatively heavier model requiring 33.5M parameters, whereas the proposed CNN is built with 25.2M parameters using the same backbone, which itself has 21.0M parameters. Thus, our method indicates competitive performance on the Sports-102 dataset.

**Dance-20**: POA-Net has achieved 76.73% accuracy on the Dance-20 dataset using MobileNet-V2 [3]. Remarkably, our method attains a significant improvement of 91.19% accuracy on this dataset compared to POA-Net. The proposed method attains more than 91.0% using all CNN backbones, implying its effectiveness.

It is evident that the proposed method attains decent and competitive performances on these benchmark datasets. We plan to reformulate image distributions of these datasets for evaluation, focusing on few-shot learning and domain adaptation techniques in the future.

*Ablation Study*

The ablation study justifying the key components of the deep learning models is two-fold. Firstly, the significance of added regions upon a backbone CNN is evaluated, excluding channel attention, and the results are provided in Table 9. ResNet-50 achieves 88.10% accuracy in this study, whereas the full model boosted the accuracy of 89.15%. The results imply that regions learn to represent local details, which is important for improving the discriminativeness of aggregated feature maps for classification. This gain is evident over the baseline performances using various base CNNs.

The next ablation study is conducted with an attention mechanism applied to the top of the backbone CNN, excluding the influence of local contextual regions. In this test, a channel-wise attention map is generated from the base output feature map using a *sigmoid* activation function. Then, the attention map is multiplied with the output base feature map to represent the final feature vector for classification. The results are given in Table 10. In this study, ResNet-50 has reached 81.88% accuracy. The results imply that channel-wise attention map refines the base output feature map and improves the baseline performance. However, the gain is not as high as compared to the included regions. It evinces that regions are effectively beneficial in learning and summarizing local details, which are further refined with channel-wise attention method. Hence, these ablation studies indicate the necessity of both modules in developing the proposed deep learning framework over CNN baseline.

The gradient-weighted class activation mapping (Grad-CAM) visualizations [53] reflect a deep model's ability to localize coarse regions within the input image for prediction. The Grad-CAM exploits the gradient information of the last convolutional layer of a CNN for making a prediction. It is widely used for class-discriminative visual explanations to justify the effectiveness of a model. The superimposed image illustrations (Fig. 5) clearly show relevant feature representations of complex body-part movements in diverse sport actions.

## Conclusion

A new dataset representing women's sports actions of 50 categories was developed for experimental scenarios when the training samples were limited. The performance has been assessed with a deep learning model incorporating four local regions and channel attention for feature representation. The method has been evaluated using four base CNNs. The dataset may contain a few training samples which is a challenge of this dataset. This dataset is made public to the scientific community for further enhancement and to explore wider applicability in the context of sports-based pose-estimation and object detection, sports analytics, etc. Moreover, experiments have been conducted on other sports and dance datasets to generalize the proposed method, which has achieved improved performance. For further analysis on the WomenSports dataset, few-shot learning, domain adaptation, and related other challenging experimental scenarios based on deep learning methods will be studied in the future.

**Acknowledgements**

This work is supported by the Cross-Disciplinary Research Framework (CDRF: C1/23/168) and New Faculty Seed Grant (NFSG/23-24) projects, and necessary computational infrastructure at the Birla Institute of Technology and Science (BITS) Pilani, Pilani Campus, Rajasthan, 333031, India. The authors would like to thank the Editor-in-Chief and anonymous reviewers for their valuable feedback to improve this work.

**Dataset Availability:** The images were collected from public websites and open-source datasets for research only. The WomenSports dataset is available at the Mendeley Data (*https://data.mendeley.com/datasets/ m39ymdv48s/3*), under a Creative Common 4.0 License.

## Supplementary Information

The train-test ratio representing different sports actions is given in Table 11.

**Table 11.** Image Distribution Summary of WomenSports Dataset. The training data is 5% and the test data is 95%. In another data distribution, 5% validation and 90% test data, keeping the same training data is followed.

| No. | Sports Class | Train | Test | Total | No. | Sports Class | Train | Test | Total |
|---|---|---|---|---|---|---|---|---|---|
| 1 | Archery | 30 | 574 | 604 | 2 | Artistic_Swimming | 30 | 553 | 583 |
| 3 | Badminton | 32 | 625 | 657 | 4 | Baseball | 30 | 571 | 601 |
| 5 | Basketball | 40 | 757 | 797 | 6 | Bobsleigh | 32 | 608 | 640 |
| 7 | Boxing | 35 | 648 | 683 | 8 | Canoeing | 33 | 646 | 679 |
| 9 | Cheerleading | 32 | 611 | 643 | 10 | Cricket | 35 | 667 | 702 |
| 11 | Curling | 30 | 583 | 613 | 12 | Cycling | 32 | 611 | 643 |
| 13 | Diving | 30 | 587 | 617 | 14 | Equestrian | 33 | 635 | 668 |
| 15 | Fencing | 33 | 646 | 679 | 16 | Football | 33 | 646 | 679 |
| 17 | Golf | 32 | 604 | 636 | 18 | Gymnastic-Beam | 30 | 583 | 613 |
| 19 | Gymnastic-Floor | 33 | 635 | 668 | 20 | Hammer-Throw | 33 | 635 | 668 |
| 21 | Handball | 33 | 629 | 662 | 22 | High-Jump | 32 | 594 | 626 |
| 23 | Hockey | 36 | 691 | 727 | 24 | Ice-Hockey | 36 | 691 | 727 |
| 25 | Javelin | 35 | 665 | 700 | 26 | Judo | 40 | 708 | 748 |
| 27 | Kabaddi | 30 | 571 | 601 | 28 | Long-Jump | 40 | 729 | 769 |
| 29 | Pole-Vault | 40 | 711 | 751 | 30 | Polo | 31 | 607 | 638 |
| 31 | Rowing | 32 | 593 | 625 | 32 | Rugby | 35 | 657 | 692 |
| 33 | Sailing | 33 | 634 | 667 | 34 | Shooting-Rifle | 32 | 592 | 624 |
| 35 | Skateboarding | 33 | 644 | 677 | 36 | Skating | 35 | 661 | 696 |
| 37 | Skiing | 33 | 638 | 671 | 38 | Snooker | 32 | 591 | 623 |
| 39 | Sprint-Run | 35 | 662 | 697 | 40 | Surfing | 33 | 629 | 662 |
| 41 | Swimming | 33 | 640 | 673 | 42 | Table-Tennis | 33 | 628 | 661 |
| 43 | Tennis | 36 | 701 | 737 | 44 | UFC | 36 | 692 | 728 |
| 45 | Volleyball | 35 | 665 | 700 | 46 | Wall-Climbing | 35 | 653 | 688 |
| 47 | Waterpolo | 33 | 634 | 667 | 48 | Weight-Lifting | 33 | 631 | 664 |
| 49 | Wrestling | 33 | 646 | 679 | 50 | WWE | 34 | 617 | 651 |

**Confusion Matrix:** The baseline confusion matrix is shown in the left and the confusion matrix of proposed method is shown at the right side for a contrasting comparison in Fig. 6. The confusion matrices of proposed method using other two base CNNs are shown in the bottom row in Figures 6 - 8.

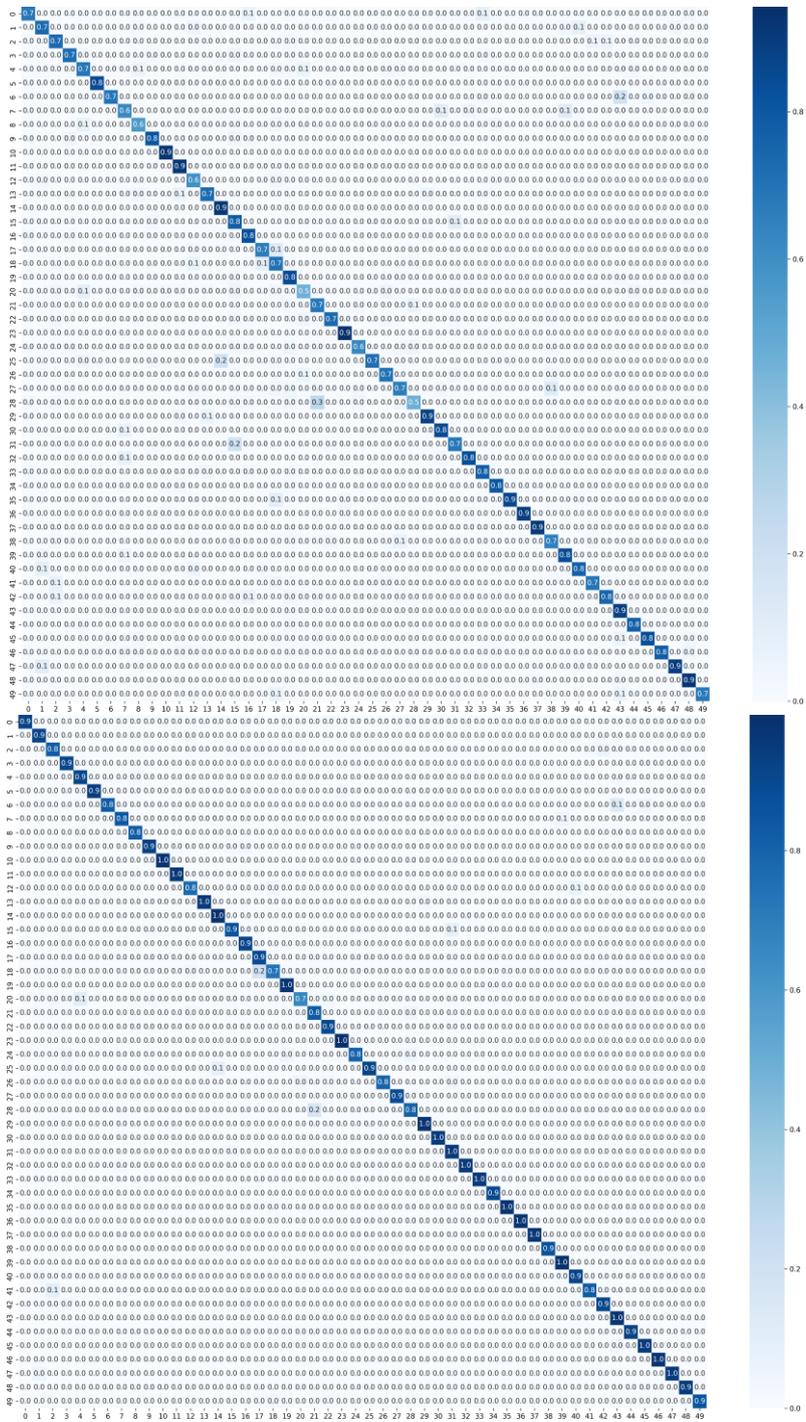

**Figure 6.** Confusion matrices of baseline (top) and proposed method (bottom) using ResNet-50 on the WomenSports dataset.

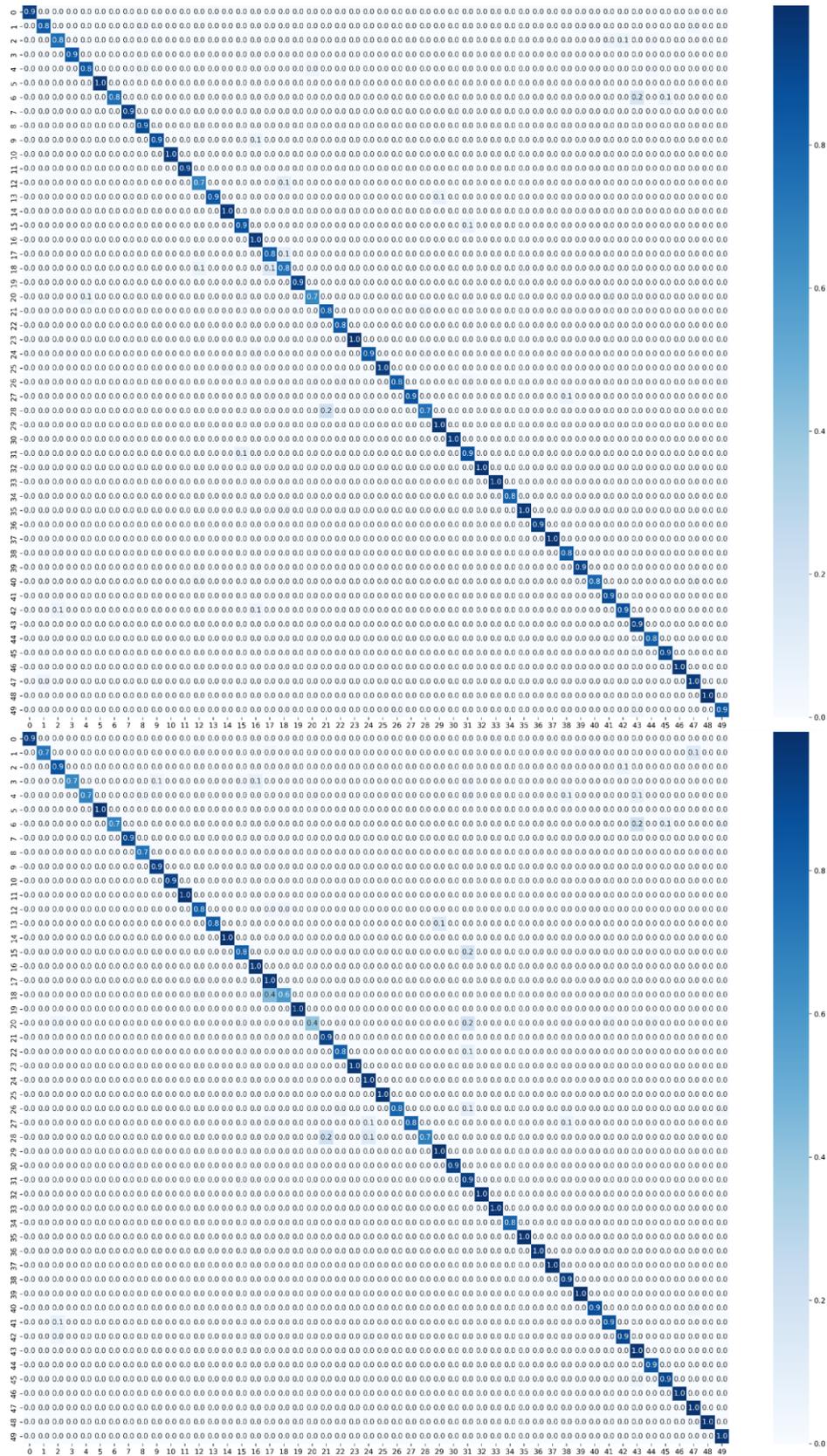

**Figure 7.** Confusion matrices of proposed method using Xception (top) and MobileNet-V2 (bottom) on the WomenSports dataset.

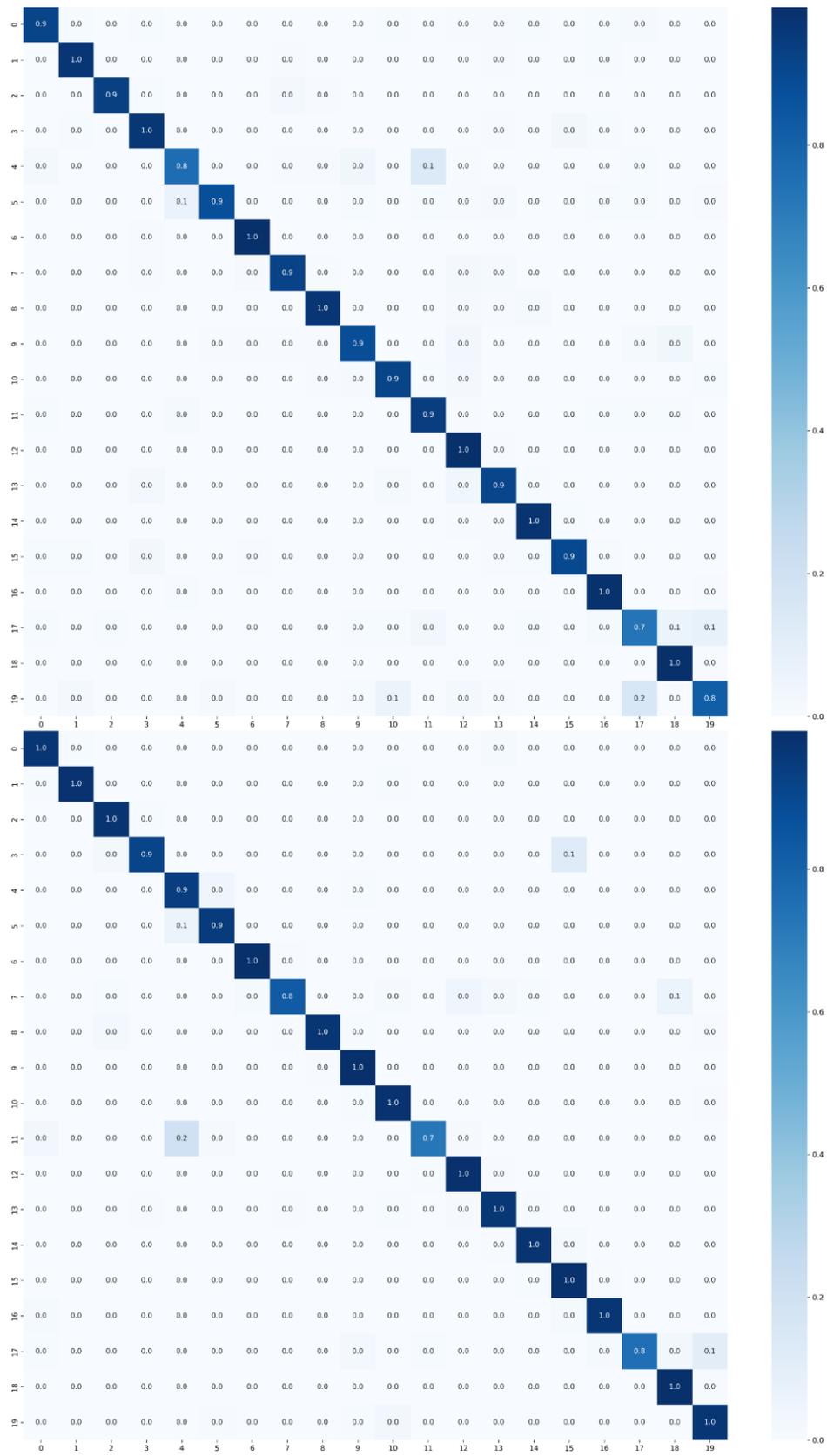

**Figure 8.** Confusion matrix of proposed method using ResNet-50 (top) MobileNet-V2 (bottom) on the Dance-20 dataset.